\documentclass[authoryear,preprint,12pt]{elsarticle}

\usepackage{epsfig}
\usepackage{booktabs}
\usepackage{amssymb}
\usepackage{multirow}
\usepackage{mathrsfs}
\usepackage{longtable}
\usepackage{lscape}
\usepackage{tabularx}
\usepackage{bm}

\newdefinition{example}{Example}
\newdefinition{definition}{Definition}

\usepackage{lineno}

\journal{European Journal of Operational Research}

\begin{document}

\begin{frontmatter}



\title{Parameter estimation based on interval-valued belief structures}

\author[swu]{Xinyang Deng}
\author[GUFS]{Yong Hu}
\author[HKPU]{Felix T. S. Chan}
\author[vu]{Sankaran Mahadevan}
\author[swu,vu]{Yong Deng\corref{cor}}
\ead{prof.deng@hotmail.com; ydeng@swu.edu.cn}
\cortext[cor]{Corresponding author.}

\address[swu]{School of Computer and Information Science, Southwest University, Chongqing, 400715, China}
\address[GUFS]{Institute of Business Intelligence and Knowledge Discovery, Guangdong University of Foreign Studies, Sun Yat-Sen University, Guangzhou, 510006, China}
\address[HKPU]{Department of Industrial and Systems Engineering, Hong Kong Polytechnic University, Hong Kong, China}
\address[vu]{School of Engineering, Vanderbilt University, Nashville, TN, 37235, USA}

\begin{abstract}
Parameter estimation based on uncertain data represented as belief structures is one of the latest problems in the Dempster-Shafer theory. In this paper, a novel method is proposed for the parameter estimation in the case where belief structures are uncertain and represented as interval-valued belief structures. Within our proposed  method, the maximization of likelihood criterion and minimization of estimated parameter's uncertainty are taken into consideration simultaneously. As an illustration, the proposed method is employed to estimate parameters for deterministic and uncertain belief structures, which demonstrates its effectiveness and versatility.
\end{abstract}
\begin{keyword}
Parameter estimation \sep Interval-valued belief structures \sep Dempster-Shafer theory \sep Maximum likelihood estimation
\end{keyword}
\end{frontmatter}

\section{Introduction}\label{Introduction}
Dempster-Shafer theory (D-S theory for short) \citep{Dempster1967, Shafer1976} has been widely used because it allows to handle uncertain data \citep{durbach2012modeling,Yang2013393,Yang20131AI}. In D-S theory, various belief structures are employed to represent the uncertain data. Recently, the study of parameter estimation based on belief structures has attracted many attentions \citep{Come2009334,denoeux2010maximumSMPS,denoeux2013maximum,Su20131217IJAR}. Typically, Denoeux \citep{denoeux2013maximum} proposed an evidential EM algorithm for parameter estimation in the case of crisp belief structures, and Su et al. \citep{Su20131217IJAR} developed a parameter estimation approach for fuzzy belief structures. In this paper, the parameter estimation based on interval-valued belief structures \citep{yager2001dempster, Wang200635} has been considered. A novel parameter estimation method is proposed for the case of interval-valued belief structures. Within the proposed method, two criteria, the maximization of observation data's likelihood and the minimization of estimated parameter's uncertainty, are both considered simultaneously. The proposed method is effective for both crisp (deterministic) and interval-valued (uncertain) belief structures, and promising for various applications.

\section{D-S theory and belief structures}\label{}
D-S theory \citep{Dempster1967, Shafer1976} is often regarded as an extension of the Bayesian theory. Please refer to \citep{Shafer1976,Yang20131AI} for more knowledge about D-S theory. In D-S theory, various belief structures, such as crisp, interval-valued and fuzzy belief structures, are employed as basic data structures. They are used to express various uncertain information. A crisp belief structure is defined as follows.
\begin{definition}
Let a finite nonempty set $\Omega$ be a frame of discernment, and $2^\Omega$ denote the power set of $\Omega$. A crisp belief structure is a mapping $m: 2^\Omega \to [0,1]$, satisfying
\begin{equation}
m(\emptyset ) = 0 \quad and \quad \sum\limits_{A \in 2^\Omega}
{m(A) = 1}
\end{equation}
\end{definition}

The crisp belief structure is deterministic because its belief degree is expressed by real numbers. By contrast, the interval-valued belief structure (IBS) is a kind of uncertain belief structures, which is an extension of the crisp belief structure. It is more capable to represent the uncertain information. Some basic concepts about IBS are given as below \citep{yager2001dempster, Wang200635}.
\begin{definition}
Let $\Omega$ be a frame of discernment, $F_1$, $F_2$, $\cdots$, $F_n$ be the $n$ focal elements on $\Omega$. An IBS $m_I$ satisfies such conditions
\begin{enumerate}[1.]
  \item $a_i \le m_I(F_i) \le b_i$, where $a_i, b_i \in [0,1]$ and $i = 1, 2, \cdots, n$;
  \item $\sum\nolimits_{i = 1}^n {a_i }  \le 1$ and $\sum\nolimits_{i = 1}^n {b_i }  \ge 1$;
  \item $m_I(F) = 0$, $\forall F \notin \{ F_1 ,F_2 , \cdots ,F_n \}$.
\end{enumerate}
\end{definition}

An IBS is valid if it satisfies $\sum\nolimits_{i = 1}^n {a_i }  \le 1$ and $\sum\nolimits_{i = 1}^n {b_i } \ge 1$. In the rest of this paper, all the IBSs are valid.

\section{Proposed parameter estimation method}
In previous literatures \citep{denoeux2013maximum,Su20131217IJAR}, parameter estimation based on crisp and fuzzy belief structures has been studied. However, the parameter estimation based on interval-valued belief structures is still an unsettled problem. In this paper, a novel parameter estimation method based on IBSs is proposed to fill that gap. Without loss of generality, some concepts about interval probabilities are introduced first.
\subsection{Interval probabilities}
\begin{definition}\citep{Guo2010444}
Let $X$ be a finite set $X = \{x_1, \cdots, x_n \}$, a set of intervals ${P_I} = \{ I_i = [w_i^-, w_i^+], i = 1, \cdots, n \}$ satisfying $0 \le w_i^- \le w_i^+ \le 1$ is an interval probabilities of $X$ if there are $w_i^* \in [w_i^-, w_i^+]$ for $i = 1, \cdots, n$ such that $\sum\limits_{i = 1}^n {w_i^* }  = 1$.
\end{definition}

Interval probabilities are the extension of point-valued probability mass functions, which can be degenerated to the classical probability distribution.
\begin{definition}\citep{Guo2010444}
Let ${P_I} = \{ I_i = [w_i^-, w_i^+], i = 1, \cdots, n \}$ be an interval probabilities, the $\alpha$th ignorance of ${P_I}$, denoted as $I^{\alpha}(P_I)$, is
\begin{equation}
I^\alpha  (P_I ) = {{\sum\limits_{i = 1}^n {(w_i^ +   - w_i^ -  )^\alpha  } } \mathord{\left/ {\vphantom {{\sum\limits_{i = 1}^n {(w_i^ +   - w_i^ -  )^\alpha  } } n}} \right. \kern-\nulldelimiterspace} n}
\end{equation}
\end{definition}

Obviously, $I^\alpha  (P_I ) \in [0, 1]$. $I^\alpha  (P_I ) = 1$ for $I_1 = I_2 = \cdots = I_n = [0, 1]$ and $I^\alpha  (P_I ) = 0$ for the point-valued probabilities. $I^1(P_I)$ can be seen as an effective index to measure the uncertainty/imprecision of interval probabilities.

\subsection{Likelihood function model for IBS}
To do the parameter estimation under IBS environment, the likelihood function model for IBS should be developed first. Let $X$ be a discrete random variable taking values in $\Omega_X = \{H_1, H_2, \cdots, H_q\}$, with interval probabilities $p_X ( \cdot ;\theta )$ which depends on unknown parameter ${\Theta} = \{ \theta_i = [\theta_i^-, \theta_i^+], i = 1, \cdots, q \}$. There are several types of observational data.

If the observational data is completely certain, for example $H_i$ happened, the likelihood function given a singleton $H_i$ can be represented as
\begin{equation}
L(H_i ;\Theta ) = [\theta _i^ -  , \quad \theta _i^ +  ]
\end{equation}

If an event $F$, $F \subseteq \Omega_X$, is observed, the likelihood function given a subset $F$  is now
\begin{equation}
L(F;\Theta ) = [L_F^ -  ,\quad L_F^ +  ]
\end{equation}
where $L_F^ -   = \max \left[ {\sum\limits_{H_i  \subseteq F} {\theta _i^ -  } ,\quad (1 - \sum\limits_{H_i  \not\subset F} {\theta _i^ + } )} \right]$, $L_F^ +   = \min \left[ {\sum\limits_{H_i  \subseteq F} {\theta _i^ +  } ,\quad (1 - \sum\limits_{H_i  \not\subset F} {\theta _i^ -  } )} \right]$

If the observational data is described by a piece of uncertain belief structure --- an IBS $m_I$, the likelihood function given such uncertain data is
\begin{equation}
L(m_I ;\Theta ) = [L_{m_I }^ -  ,\quad L_{m_I }^ +  ]
\end{equation}
where
\begin{equation}
\begin{array}{l}
 L_{m_I }^ - / L_{m_I }^ +   = \min / \max \quad \sum\limits_{i = 1}^n {m_I (F_i )L_{F_i }^* }  \\
  \quad \quad s.t.\quad \sum\limits_{i = 1}^n {m_I (F_i )}  = 1 \\
  \quad \quad \quad \quad a_i  \le m_I (F_i ) \le b_i ,\quad \forall i = 1, \cdots ,n \\
  \quad \quad \quad \quad L_{F_i }^ -   \le L_{F_i }^*  \le L_{F_i }^ +  ,\quad \forall i = 1, \cdots ,n \\
 \end{array}
\end{equation}

Now assuming there are $p$ observational data, expressed by $p$ IBSs, $\bm{m_I} = (m_{I_1}, m_{I_2}, \cdots, m_{I_p})$. The likelihood of $\bm{m_I}$ is represented as
\begin{equation}
L(\bm{m_I} ;\Theta ) = [L_{\bm{m_I} }^ -  ,\quad L_{\bm{m_I} }^ +  ] = [\prod\limits_{i = 1}^p {L_{m_{I_i } }^ -  } ,\quad \prod\limits_{i = 1}^p {L_{m_{I_i } }^ +  } ]
\end{equation}
\setlength{\abovecaptionskip}{-10pt}
\setlength{\belowcaptionskip}{-10pt}
\begin{table}\footnotesize
    \caption{Observational data represented as crisp belief structures}\label{Deterministic}
    \begin{center}
    \begin{tabular}{lcccccc}
    \toprule
    Observation &  1 & 2 & 3 & 4 & 5 & 6  \\
    \midrule
    $m(\{a\})  $  &   1.0 & 1.0 & 1.0 & 0.3 & 0.0 & 0.0  \\
    $m(\{b\})  $  &   0.0 & 0.0 & 0.0 & 0.3 & 1.0 & 1.0  \\
    $m(\{a,b\})$  &   0.0 & 0.0 & 0.0 & 0.4 & 0.0 & 0.0  \\
    \bottomrule
    \end{tabular}
    \end{center}
\end{table}

\subsection{Solution for parameter estimation}
The likelihood function model developed above is the foundation for the parameter estimation based on IBSs. Depending on that, an optimization model $P$ is proposed to make an estimation for parameter $\Theta$.
\begin{equation}
Model \; P: \quad \mathop {\arg \max }\limits_\Theta  \quad D \left( {L(\bm{m_I} ;\Theta )}, \; [0,0] \right) - I^{\alpha} \left( \Theta \right)
\end{equation}
where $I^{\alpha} \left( \Theta \right)$ is the $\alpha$th ignorance of $\Theta$, and $D \left( {L(\bm{m_I} ;\Theta )}, \; [0,0] \right)$ is a distance measure for two intervals ${L(\bm{m_I} ;\Theta )}$ and $[0,0]$ presented in \citep{tran2002comparison}. For $A = [a^-, a^+]$ and $B = [b^-, b^+]$,
\begin{equation}
\begin{array}{l}
D(A,B) = \sqrt {\left[ {\left( {\frac{{a^ -   + a^ +  }}{2}} \right) - \left( {\frac{{b^ -   + b^ +  }}{2}} \right)} \right]^2  + \frac{1}{3}\left[ {\left( {\frac{{a^ +   - a^ -  }}{2}} \right)^2  + \left( {\frac{{b^ +   - b^ -  }}{2}} \right)^2 } \right]}  \\
\end{array}
\end{equation}

The model $P$ is formulated based on two criteria, namely maximization of the likelihood of observational data indicated by $D \left({L(\bm{m_I} ;\Theta )}, \; [0,0] \right)$ and minimization of the uncertainty/imprecision of estimated parameter indicated by $I^{\alpha} \left( \Theta \right)$, respectively. Within model $P$, $\alpha$ is a control parameter to adjust the impact of these two criteria. A point-valued probability distribution will be obtained if $\alpha = 1$, and a set of interval probabilities will be obtained if $\alpha \ge 2$. A global optimization algorithm called OQNLP \citep{ugray2007scatter} is employed to solve the optimization model $P$. Generally, the proposed method is superior to Denoeux's method because not only can this method handle the case of crisp belief structures, but it can also deal with the case of IBSs, as shown in the next section.
\begin{table}\footnotesize
    \caption{Results of parameter estimation for the case of crisp belief structures}\label{ResultsEx1}
    \begin{center}
    \begin{tabular}{lcccc}
    \toprule
    Probability  & & $p(a)$ & & $p(b)$  \\
    \midrule
    Denoeux's method \citep{denoeux2013maximum} & & 0.6 & & 0.4  \\
    Proposed method ($\alpha = 1$) & & 0.6 & & 0.4  \\
    \bottomrule
    \end{tabular}
    \end{center}
\end{table}
\begin{table}\footnotesize
    \caption{Observational data represented as IBSs}\label{IBSsData}
    \begin{center}
    \begin{tabular}{lcccc}
    \toprule
    Observation &  1 & 2 & 3 & 4  \\
    \midrule
    $m_I(\{ H_1 \})$  &           [0.30, 0.40] & [0.35, 0.45] & [0.10, 0.25] & [0.30, 0.45]  \\
    $m_I(\{ H_2 \})$  &           [0.10, 0.25] & [0.10, 0.20] & [0.30, 0.45] & [0.30, 0.50]  \\
    $m_I(\{ H_3 \})$  &           [0.25, 0.35] & [0.20, 0.30] & [0.35, 0.50] & [0.15, 0.40]  \\
    $m_I(\{ H_1, H_2, H_3 \})$  & [0.10, 0.20] & [0.05, 0.15] & [0.10, 0.25] & [0.00, 0.20]  \\
    \bottomrule
    \end{tabular}
    \end{center}
\end{table}
\begin{table}\footnotesize
    \caption{Results of parameter estimation for the case of IBSs}\label{ResultsEx2}
    \begin{center}
    \begin{tabular}{lcccc}
    \toprule
    $\alpha$'s value   & $P_I(H_1)$ & $P_I(H_2)$ & $P_I(H_3)$ & $I^1(P_I)$\\
    \midrule
    $\alpha = 1$  & [0.9823, 0.9823] & [0.0000, 0.0000] & [0.0177, 0.0177] &  0.0000 \\
    $\alpha = 2$  & [0.8397, 0.9433] & [0.0057, 0.1093] & [0.0510, 0.1547]  & 0.1036 \\
    $\alpha = 3$  & [0.5821, 0.9331] & [0.0122, 0.3632] & [0.0547, 0.4058]  & 0.3510\\
    $\alpha = 4$  & [0.4614, 0.9569] & [0.0085, 0.5040] & [0.0346, 0.5301] &  0.4955\\
    $\alpha = 5$  & [0.2963, 0.8751] & [0.0324, 0.6112] & [0.0925, 0.6713] &  0.5788\\
    $\alpha = 6$  & [0.2580, 0.8907] & [0.0288, 0.6615] & [0.0805, 0.7132] &  0.6327\\
    $\alpha = 7$  & [0.3228, 0.9988] & [0.0002, 0.6763] & [0.0010, 0.6770]  & 0.6760\\
    $\alpha = 8$  & [0.2687, 0.9744] & [0.0055, 0.7112] & [0.0201, 0.7259]  & 0.7057\\
    $\alpha = 9$  & [0.1876, 0.9136] & [0.0235, 0.7496] & [0.0629, 0.7889]  & 0.7260\\
    $\alpha = 10$  & [0.2272, 0.9768] & [0.0050, 0.7546] & [0.0182, 0.7678] &  0.7496\\
    $\alpha = 20$  & [0.1339, 0.9815] & [0.0042, 0.8518] & [0.0143, 0.8619]  & 0.8476\\
    \bottomrule
    \end{tabular}
    \end{center}
\end{table}

\section{Numerical Examples}
\subsection{Example for the case of crisp belief structures}
In this example, a set of observational data is composed of 6 crisp belief structures, as shown in Table \ref{Deterministic}. The estimated parameter is the probability distribution of random variable $X$ taking values in $\Omega_X = \{a, b\}$. Two methods, namely Denoeux's \citep{denoeux2013maximum} and proposed in this paper, are employed. As shown in Table \ref{ResultsEx1}, the results obtained by these two methods are identical, which demonstrates the proposed method is effective for crisp belief structures.

\subsection{Example for the case of interval-based belief structures}
While, Denoeux's method is incapable when the observational data are represented by IBSs, as shown in Table \ref{IBSsData}. In this situation, a set of interval probabilities can be estimated based on different $\alpha$ by using the proposed method. As seen in Table \ref{ResultsEx2}, a point-valued probability distribution is obtained that $p(H_1) = 0.9823$, $p(H_2) = 0.0$, $p(H_3) = 0.0177$ when $\alpha = 1$. The estimated probability distribution becomes a set of interval probabilities when $\alpha \ge 2$. The uncertainty of obtained interval probabilities rises with the increase of $\alpha$.

\setlength{\abovecaptionskip}{-10pt}
\setlength{\belowcaptionskip}{-10pt}
\begin{table}\scriptsize
    \caption{The HIS trustworthiness evaluation on each criterion}\label{DataEx3}
    \begin{center}
    \begin{tabular}{ll}
    \toprule
    Criteria  &  HIS trustworthiness evaluations  \\
    \midrule
    Reliability   & \{($A$,[0.0393,0.2159]), ($G$,[0.3305,0.6476]), ($V$,[0.2266,0.5128]), ($E$,[0,0.1026]), ($\Omega$,[0,0.1449])\} \\
    Safety  &   \{($G$,[0.0728,0.3119]), ($V$,[0.4817,0.8246]), ($E$,[0.068,0.1832]), ($\Omega$,[0,0.1814])\} \\
    Real-time   &  \{($V$,[0.229,0.7]), ($E$,[0.2727,0.75]), ($\Omega$,[0,0.1778])\}  \\
    Maintainability  &  \{($G$,[0.1515,0.2849]), ($V$,[0.4545,0.6648]), ($E$,[0.048,0.2424]), ($\Omega$,[0,0.186])\} \\
    Availability    &  \{($A$,[0.0867,0.2537]), ($G$,[0.5034,0.7258]), ($V$,[0.1438,0.2722])($\Omega$,[0,0.0899])\} \\
    Security  &   \{($G$,[0.0513,0.1967]), ($V$,[0.3213,0.473]), ($E$,[0.4017,0.5676]), ($\Omega$,[0,0.0939])\} \\
    \bottomrule
    \end{tabular}
    \end{center}
\end{table}

\subsection{Trustworthiness assessment of hospital information system (HIS)}
In this example, the trustworthiness assessment of HIS is studied. Generally, the rating of HIS trustworthiness can be $\Omega = \{Poor, Average, Good, VeryGood, Excellent\}$. Table \ref{DataEx3} shows the assessment criteria and the evaluation for each criterion in a HIS, derived from literature \citep{fu2012conjunctive}. Here, the evaluations on various criteria are treated as a set of observational data composed of IBSs. Based on our proposed method, various sets of interval probabilities are obtained when $\alpha$ takes different values, as shown in Table \ref{ResultsEx3}. According to these results, the rating $VeryGood$ is appropriate for the given HIS.
\begin{table}\scriptsize
    \caption{Results of parameter estimation for HIS trustworthiness assessment}\label{ResultsEx3}
    \begin{center}
    \begin{tabular}{lccccc}
    \toprule
    $\alpha$'s value   & $P_I(Poor)$ & $P_I(Average)$ & $P_I(Good)$ & $P_I(VeryGood)$ & $P_1(Excellent)$\\
    \midrule
    $\alpha = 1$  & [0.0000,    0.0000]  &  [0.0000,    0.0000]  &  [0.0000,    0.0000]  &  [1.0000,    1.0000]  &  [0.0000,    0.0000] \\
    $\alpha = 2$  & [0.0007,    0.0291]  &  [0.0007,    0.0558]  &  [0.0015,    0.1789]  &  [0.8187,    0.9961]  &  [0.0010,    0.1784] \\
    $\alpha = 3$  & [0.0004,    0.0874]  &  [0.0004,    0.2284]  &  [0.0008,    0.4799]  &  [0.5187,    0.9978]  &  [0.0006,    0.4796]	\\
    $\alpha = 4$  & [0.0001,    0.0994]  &  [0.0001,    0.3615]  &  [0.0001,    0.6076]  &  [0.3919,    0.9996]  &  [0.0001,    0.6075]	\\
    $\alpha = 5$  & [0.0003,    0.1422]  &  [0.0003,    0.4503]  &  [0.0005,    0.6729]  &  [0.3262,    0.9986]  &  [0.0004,    0.6727]	\\		
    \bottomrule
    \end{tabular}
    \end{center}
\end{table}

\section{Conclusion}
In this paper, the problem of parameter estimation based on belief structures has been studied. The proposed method provides a unified framework for this problem. Not only crisp belief structures but also uncertain belief structures --- IBSs, are both can be handled. As an important technique in D-S theory, it has the ability to handle various types of uncertain data and knowledge represented as belief structures.

\section*{Acknowledgements}
The work is supported by National Natural Science Foundation of China, Grant No. 61174022.

\bibliographystyle{elsarticle-harv}
\bibliography{References}







\end{document}